\def\year{2020}\relax
\documentclass[letterpaper]{article} 
\usepackage{aaai20}  
\usepackage{times}  
\usepackage{helvet} 
\usepackage{courier}  
\usepackage[hyphens]{url}  
\usepackage{graphicx} 
\usepackage{graphicx}  
\title{Ruminating Word Representations with Random Noised Masker}

\usepackage{url}
\usepackage{boldline}
\usepackage{multirow}
\usepackage{graphicx}
\usepackage{amsmath}
\usepackage{algorithm}
\usepackage{algorithmic}

\newcolumntype{C}[1]{>{\centering\arraybackslash}m{#1}}

\author{Hwiyeol Jo and Byoung-Tak Zhang\\
\Large\textbf{Seoul National University} \\
hwiyeolj@gmail.com, btzhang@bi.snu.ac.kr 
}

\begin{document}

\maketitle

\begin{abstract}
    We introduce a training method for both better word representation and performance, which we call \textbf{GROVER} (\textbf{G}radual \textbf{R}umination \textbf{O}n the \textbf{V}ector with mask\textbf{ER}s). The method is to gradually and iteratively add random noises to word embeddings while training a model. GROVER first starts from conventional training process, and then extracts the fine-tuned representations. Next, we gradually add random noises to the word representations and repeat the training process from scratch, but initialize with the noised word representations. Through the re-training process, we can mitigate some noises to be compensated and utilize other noises to learn better representations. As a result, we can get word representations further fine-tuned and specialized on the task. When we experiment with our method on 5 text classification datasets, our method improves model performances on most of the datasets. Moreover, we show that our method can be combined with other regularization techniques, further improving the model performance.
\end{abstract}

\section{Introduction}
\subsection{Motivation}
    Most of the machine learning methodology can be defined as getting computational representations from real-life objects and then classifying the representations according to their tasks. Therefore, there have been two main approaches to increase the model performance: (1) starting with better representations from data~\cite{melamud2016context2vec,peters2018deep}, and (2) building more complex architectures that are able to extract important features and generate higher level representations~\cite{vaswani2017attention,conneau2017very}.\\
    For better initial representations, many NLP researchers have used pretrained word vectors, trained on substantially large corpus through unsupervised algorithms like word2vec~\cite{mikolov2013efficient}, GloVe~\cite{pennington2014glove}, and fastText~\cite{bojanowski2016enriching}. The pretrained word vectors not only represent the general meaning of words but also increase the model performances on most of NLP tasks~\cite{turian2010word}. Further, word vector post-processing researches~\cite{faruqui2015retrofitting,vulic2017cross,mrkvsic2017semantic,jo2018extrofitting} have tried to enrich the pretrained representations using external semantic lexicons. The post-processing methods compensate for the weak point of word vector generation algorithms, which highly depend on word order, and increases the model performance.\\
    When training an NLP model, we first initialize word representations with pretrained word vectors and then update both the model parameters and the word representations together. However, the model performance can be limited due to the initial word vectors. 
    The pretrained word representations have the general meaning of words, but the words in some tasks do not have the meaning. Although the different meanings can be learned through the training process, it could fail to learn in the context. Since the word vectors are updated through gradient descent algorithm, the values are changed slightly, and the word vectors are easy to converge on local minima.\\
    Therefore, we propose a simple trick to find better representations by using random noise maskers on the word vectors during iterative training process, which we call \textbf{GROVER} (\textbf{G}radual \textbf{R}umination \textbf{O}n the \textbf{V}ector with mask\textbf{ER}s). We expect that the noises help the model learn better representation. Additionally, GROVER can regularize the model by adding random noises to the fine-tuned word vectors. We show that proper degree of noises not only help a model learn task-specific representations but also regularize the model, improving the model performance.

\subsection{Contribution}
    Our contributions through \textbf{GROVER} are summarized as follows:
    \begin{itemize}
    \item GROVER mitigates word vector updating problem and further fine-tunes the word vectors on the task.
    \item GROVER increases the model performance by further fine-tuning word representations. Also, Our method can be applied to any model architectures using word embeddings.
    \item GROVER regularizes the model and can be combined with other regularization techniques, further increasing the model performance.
    \end{itemize}
    
\section{Related Works}
\subsection{Word-level Noises}
    Adding noises to input data is an old idea~\cite{plaut1986experiments}. However, there are few research on word-level noises since the small noises on language can change the meaning of the words.\\
    {\bf Word Dropping.} NLP tasks which utilize the text as a form of sentence and phrase consider each word as features. However, lots of features can lead a model to be overfitted to the training data due to the curse of dimensionality. Therefore, the easiest way to reduce the number of features is to drop words in the sentence at random.\\
    {\bf Word Embedding Perturbation.} \citeauthor{miyato2016adversarial} tried to use word embedding perturbation for model regularization through adversarial training framework~\cite{miyato2016adversarial}. \citeauthor{cheng2018towards} utilized the noises to build robust machine translation model~\cite{cheng2018towards}. Also, there was an approach that considers the perturbation as data augmentation~\cite{zhang2018word}.
    The previous works added the noises to all word embeddings when they are used. So, the methods can regularize the models, particularly model weights, but do not care word representations. However, our method gradually adds noises to word embeddings according to word frequency. Also, we use iterative training process to benefit from the noises and get better word representations. 

\subsection{Regularization Techniques}
    Some research already explained that the normalization can be used to regularize models~\cite{van2017l2,luo2018towards,hoffer2018norm}.\\
    {\bf Dropout}~\cite{srivastava2014dropout} is applied to neural network models, masking random neurons with 0. Dropout randomly and temporarily removes the activations during training, so the masked weights are prevented from updating. As a result, the model is prevented from over-tuning on specific features, which brings regularization. Also, dropout discourses the weights to coadapt and carries out ensemble effects to the model.\\
    {\bf Batch Normalization (BN)}~\cite{ioffe2015batch} normalizes the features according to mini-batch statistics. Batch normalization enables the features to avoid covariate shift--the weight gradients are highly dependent on the gradients of previous layers. Besides, batch normalization speeds up the training process by reshaping loss function.\\
    {\bf Layer Normalization (LN)}~\cite{ba2016layer} also utilizes mini-batch statistics to normalize the features. The difference with batch normalization is that layer normalization normalizes the inputs across the features. The statistics are computed across each feature, which is the same for all feature dimensions.

    \begin{figure*}[ht!] \centering
    \includegraphics[scale=0.32]{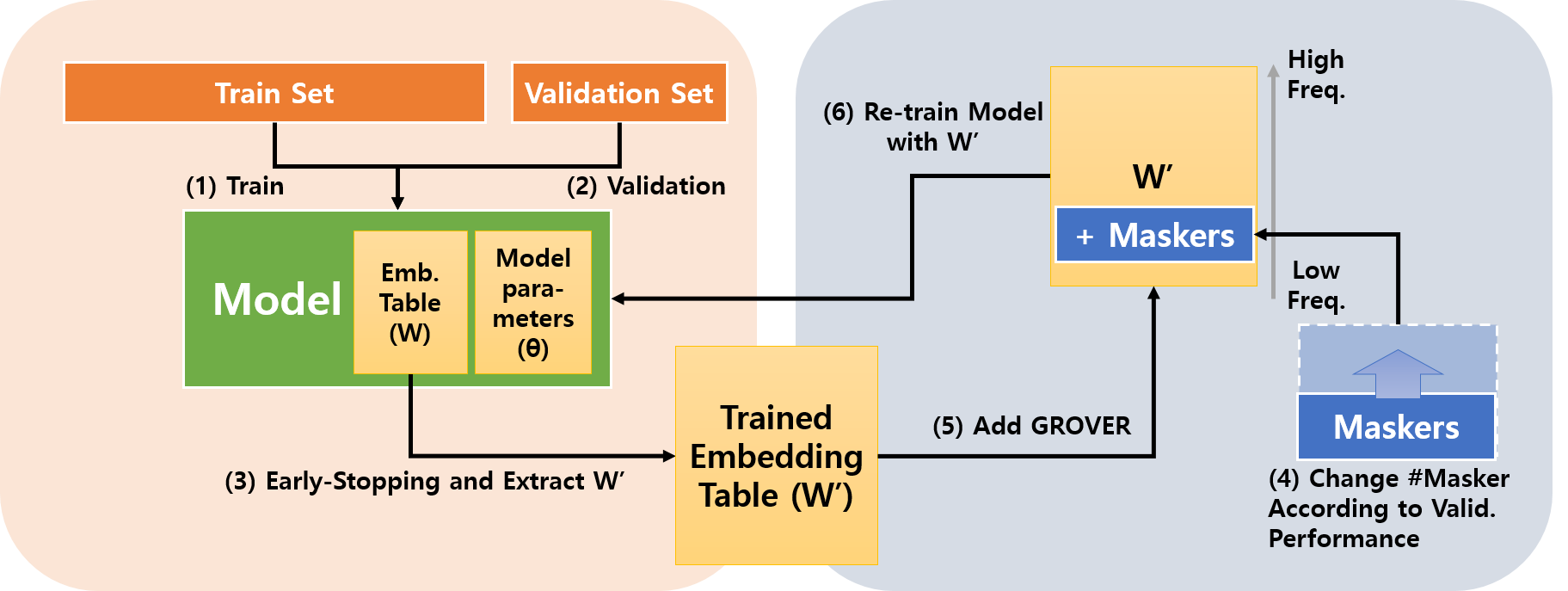}
    \caption{The flow of training framework with GROVER. (1) We first train a classifier using training set, (2-3) do early-stopping using validation set, (3) extract fine-tuned word embedding table, (4) change the number of random maskers according to validation performance, (5) gradually add the maskers to the word embeddings from low frequency words to high frequency words, and lastly (6) train the classifier with $W'$ from the very first step. We repeat the process until all the words are masked.}
    \label{fig:1}
    \end{figure*}

\subsection{Pretrained Representations}
    The representations fine-tuned by GROVER can be regarded as pretrained representations from the previous training process.\\
    {\bf Pretrained Embedding Vector} is also called pretrained word representation. According to distributional representation hypothesis~\cite{mikolov2013distributed}, pretrained embedding vectors are composed of pairs of (token, n-dimensional float vector). The word vectors usually are learned by unsupervised algorithms (e.g., word2vec~\cite{mikolov2013efficient}, GloVe~\cite{pennington2014glove}, fastText~\cite{bojanowski2016enriching}) on substantial corpus to represent general meanings of words. The pretrained embedding vectors are widely used to initialize the word vectors in models.\\
    {\bf Pretrained Embedding Model} is suggested recently to get a deep representation of each word in the context. Previous research~\cite{mccann2017learned,peters2018deep,devlin2018bert} trained deep architecture models and then utilized the model weights to represent words by using the outputs of models.\\
    We take pretrained embedding vector approach because we believe that the contextual meaning of the words can be learned through training processes if the model architecture is built to process the sequential information of the words.
    
\section{Proposed Method}
\subsection{Overall Process}
    The overall process including GROVER is illustrated in Figure~\ref{fig:1}. GROVER is applied to conventional training framework with early-stopping, but it needs a meta-level approach that trains the model again. The iterative training process will be denoted as a meta-epoch.\\
    When a training process finishes, we extract the fine-tuned word embeddings ($W'$). Next, we add maskers filled with random values to $W'$, and then re-train the model from scratch with the noised word embeddings. In other words, GROVER is a process that adds random noises to fine-tuned word embeddings and then repeats the training process with the word embeddings. Observing the validation performance in every meta-epoch, we select the model and the fine-tuned embedding. The additional details in GROVER will be described.

\subsection{GROVER Details}\label{sec:3.2}
    {\bf Maskers start from infrequently used words and move to frequently used words.}
    The random noises in a portion of word embeddings change the distribution of all the word representations during the training process, since high-level representations such as sentence representation are generated from word representations. Likewise, the model performance can be affected too much by the randomly noised maskers. In order to gradually change the distribution of word representations, we produce the noises incrementally by maskers starting from infrequently used words and moving to frequently used words.\\
    {\bf Gradualness.}
    While following the aforementioned process, we change the number of random maskers according to validation performance. If the validation performance of re-trained model increases, we move the maskers to next frequently used words. Otherwise, we roll back the word embeddings to the previous one, and gradually increase the number of random maskers without moving them so that the maskers make noises on the words again in the next step. As a result, GROVER can process both words in the previous step and words in the current step. This gradualness let the word vectors benefit from noises again, and makes GROVER dynamic.\\
    {\bf Degree of Noises (Step Size \& Noise Range).}\\
    Moderate noises are required to change word vector distribution. Therefore, we have two hyperparameters related to the noises: step size and noise range.\\
    Step size is how much the random maskers move to next frequently used words in the next training process, and we set step size to 10\% of the vocabulary size as default. For example, bottom 10\% of frequency-ordered vocabulary are masked at the first training process, and then bottom 10\% to 20\% words are masked at the next step. However, note that gradualness can change the number of maskers.\\
    The maskers are filled with the random values sampled from a uniform distribution, which range is defined as noise range. Default noise range is between -1 and 1. The noise can be extended to well-defined perturbation methods like Gaussian kernels~\cite{vilnis2014word}. In ablation studies, the effects of the hyperparameters will be presented.\\
    Overall algorithms of GROVER are summarized as follows:
    \begin{algorithm}
    \caption{Training Framework with GROVER}
    \begin{algorithmic}
    \STATE \small Train set ({\bf Train}), Validation set ({\bf Val}), A classifier ($M$),\\Word embeddings ($W$), Words frequency
    \STATE - Train $M$ with $W$
    \STATE - Get trained $M'$ and fine-tuned $W'$
    \STATE - Meta-level validation; $\textbf{Acc} \leftarrow M'(\textbf{Val}; W')$
    \IF{$\textbf{MaxAcc} < \textbf{Acc}$}
        \STATE $\textbf{MaxAcc} \leftarrow \textbf{Acc}$
        \STATE $\textbf{MaskWords} \leftarrow \textbf{NextLowFreqWords}$
        \STATE $W \leftarrow W'$
    \ELSE
        \STATE $\textbf{MaskWords}$
        \STATE $\leftarrow \textbf{MaskWords} + \textbf{NextLowFreqWords}$
    \ENDIF
    \STATE $W[\textbf{\small MaskWords}] \leftarrow W[\textbf{\small MaskWords}] + U(-1,1)$
    \STATE - Repeat the training process until all the words are masked.
    \end{algorithmic}
    \end{algorithm}
    
    \begin{table*}[t] \centering
    \caption{The data information used in text classification. YahooAnswer dataset is used for 2 different tasks, which are to classify upper-level categories and to classify lower-level categories, respectively. The vocabulary size can be slightly different due to the predefined special tokens such as {\tt none} and {\tt out-of-vocabulary}}\label{tab:1}
    \begin{tabular}{|l||C{1.5cm}|C{1.5cm}|C{1.5cm}|C{1.5cm}|C{1.5cm}|C{1.5cm}|}
    \hlineB{3}
        & \bf DBpedia & \bf \begin{tabular}{@{}c@{}} {\small YahooAns.} \\ {\small (Upper) } \end{tabular} & \bf \begin{tabular}{@{}c@{}} {\small YahooAns.} \\ {\small (Lower) } \end{tabular} & \bf AGNews & \bf Yelp Reviews & \bf IMDB\\
    \hlineB{3}
    {\small\tt \#Train} & \small 560,000 & \small 133,703 & \small 133,703 & \small 120,000 & \small 650,000 & \small 25,000 \\
    \hline
    {\small\tt \#Test} & \small 70,000 & \small 23,595 & \small 23,595 & \small 7,600 & \small 50,000 & \small 25,000 \\
    \hline
    {\small\tt \#Class} & \small 14 & \small 17 & \small 280 & \small 4 & \small 5 & \small 2 \\
    \hline
    {\small\tt \#Vocab} & \small 626,717 & \small 154,142 & \small 154,142 & \small 66,049 & \small 198,625 & \small 47,113 \\
    \hlineB{3}
    \end{tabular}
    \end{table*}
    
    \begin{table*}[t] \centering
    \caption{The performance of TextCNN classifiers with GROVER used with different pretrained word vector. We use extracted token embedding of BERT, but we cannot run GROVER because of out-of-memory (OOM) in DBpedia.}\label{tab:2}
    \begin{tabular}{|l||C{1.5cm}|C{1.5cm}|C{1.5cm}|C{1.5cm}|C{1.5cm}|C{1.5cm}|}
    \hlineB{3}
        & \bf DBpedia & \bf \begin{tabular}{@{}c@{}} {\small YahooAns.} \\ {\small (Upper) } \end{tabular} & \bf \begin{tabular}{@{}c@{}} {\small YahooAns.} \\ {\small (Lower) } \end{tabular} & \bf AGNews & \bf Yelp Reviews & \bf IMDB\\
    \hlineB{3}
    {\small\tt Clf w/ Rand} & 98.00 & 65.88 & 42.73 & 89.00 & 56.56 & 76.20 \\
    \hline
    {\small\tt \ + GROVER} & +0.50 & +6.25 & +5.63 & +2.17 & +0.84 & +3.52 \\
    \hlineB{3}
    {\small\tt Clf w/ word2vec} & 98.22 & 70.71 & 42.76 & 91.20 & 56.40 & 75.87 \\
    \hline
    {\small\tt \ + GROVER} & +0.24 & +1.22 & +5.39 & +0.05 & +0.27 & +2.05 \\
    \hlineB{3}
    {\small\tt Clf w/ GloVe} & 98.70 & 74.28 & 51.49 & 91.55 & 56.81 & 81.43 \\
    \hline
    {\small\tt \ + GROVER} & -0.03 & +0.94 & +1.60 & +0.12 & +0.32 & -0.55 \\
    \hlineB{3}
    {\small\tt Clf w/ fastText} & 98.01 & 64.06 & 40.07 & 86.12 & 55.03 & 64.38 \\
    \hline
    {\small\tt \ + GROVER} & +0.45 & +7.58 & +7.94 & +4.39 & +1.07 & +12.21 \\
    \hlineB{3}
    {\small\tt Clf w/ BERTTokenEmb} & 97.81 & 55.15 & 32.74 & 89.92 & 54.53 & 64.66 \\
    \hline
    {\small\tt \ + GROVER} & OOM* & +16.35 & +13.95 & +0.94 & +2.01 & +11.03 \\
    \hlineB{3}
    {\small\tt Clf w/ ExtroGloVe} & 98.67 & 74.52 & 52.02 & 91.63 & 57.82 & 81.71 \\
    \hline
    {\small\tt \ + GROVER} & +0.05 & +1.17 & +1.80 & +0.28 & +0.16 & -0.72\\
    \hlineB{3}
    \end{tabular}
    \end{table*}

    \noindent GROVER is applied to the embeddings by adding random noises to the fine-tuned word vectors. So our method is independent of model architectures in that most NLP models use word-level embeddings. The random noises might disturb the representation, but during the re-training processes, some noises which harm the performance are compensated. On the other hand, other noises are utilized to jump over the initial word vector values if the degree of noises is moderate.\\
    The noises by GROVER prevent the model from overfitting to the validation set in re-training processes, whereas the model with GROVER incrementally fits to the validation set through early-stopping in each training process. Therefore, the model keeps fitting to the validation set with regularization, so the model performance on the test set will increase if the model performance on the validation set is correlated to the performance on the test set.

\section{Experiment}
\subsection{Datasets}
    We prepare 3 topic classification datasets; {\bf DBpedia ontology}~\cite{lehmann2015dbpedia}, {\bf YahooAnswers}~\cite{chang2008importance}, {\bf AGNews} and 2 sentiment classification datasets; {\bf Yelp reviews}~\cite{zhang2015character}, {\bf IMDB}~\cite{maas2011learning}.
    YahooAnswer dataset is used for 2 different tasks, which are to classify upper-level categories and to classify lower-level categories, respectively. The data information is presented in Table~\ref{tab:1}. We split 15\% of each train set to validation set, and each dataset has its own test set. The validation set is used for early-stopping both at every epoch and every meta-epoch. We use the first 100 words as inputs, including all special symbols in 300 dimensional embedding space.
    
\subsection{Classifier}
    We use TextCNN~\cite{kim2014convolutional} classifiers. The model consists of 2 convolutional layers with the 32 channels and 16 channels, respectively. We adopt multiple sizes of kernels--2, 3, 4, and 5, followed by ReLU activation~\cite{hahnloser2000digital} and max-pooled. We concatenate the kernels after every max-pooling layer. We optimize the model using Adam~\cite{kingma2014adam} with 1e-3 learning rate, and do early-stopping.\\
    Initial word embeddings are \textbf{GloVe}~\cite{pennington2014glove} post-processed by \textbf{extrofitting}~\cite{jo2018extrofitting} if we do not mention explicitly. We will also present the performance of GROVER on other major pretrained word embeddings.
\subsection{Regularization Implementation}
    We implement 5 regularization (including normalization) methods to compare the techniques with ours. First, {\bf word dropping} is implemented in the pre-processing part, which removes random words in the text. Reducing the number of words in the training data, word dropping results in regularization. We set the random probability $p$ as 0.1. {\bf Dropout}~\cite{srivastava2014dropout} is added to the final fully connected layer with dropout probability 0.1, which performs the best in our experiments. {\bf Batch Normalization}~\cite{ioffe2015batch} is located between each convolutional layer and an activation function, as used in the original paper. {\bf Layer Normalization}~\cite{ba2016layer} is implemented in the same position. We report the performance averaged over 10 runs.
    
    \begin{table*}[t!] \centering
    \caption{The performance of TextCNN classifiers with different regularization techniques. We also observe that GROVER positively matches with the classifier and with other regularization techniques.}\label{tab:3}
    \begin{tabular}{|l||C{1.6cm}|C{1.6cm}|C{1.6cm}|C{1.6cm}|C{1.6cm}|C{1.6cm}|}
    \hlineB{3}
        & \bf DBpedia & \bf \begin{tabular}{@{}c@{}} {\small YahooAns.} \\ {\small (Upper) } \end{tabular} & \bf \begin{tabular}{@{}c@{}} {\small YahooAns.} \\ {\small (Lower) } \end{tabular} & \bf AGNews & \bf Yelp Review & \bf IMDB\\
    \hlineB{3}
    {\small\tt Base TextCNN} & 98.67 & 74.52 & 52.02 & 91.63 & 57.82 & \bf 81.71 \\
    \hlineB{3}
    {\small\tt + WordDrop} & -0.05 & -4.17 & -8.73 & -0.50 & -1.60 & -0.88 \\
    \hline
    {\small\tt + DO (p=0.1)\ \ } & +0.05 & -0.03 & -0.03 & +0.21 & \bf +0.18 & -0.29 \\
    \hline
    {\small\tt + BN} & -0.17 & -0.10 & -0.66 & -0.22 & -0.23 & -0.44 \\
    \hline
    {\small\tt + LN} & \bf +0.07 & +0.94 & +1.35 & -0.07 & -0.19 & -0.52 \\
    \hline
    {\small\tt + GROVER} & +0.05 & \bf +1.17 & \bf +1.80 & \bf +0.28 & +0.16 & -0.72 \\
    \hlineB{3}
    \end{tabular}
    
    \qquad
    
    \begin{tabular}{|l||C{1.6cm}|C{1.6cm}|C{1.6cm}|C{1.6cm}|C{1.6cm}|C{1.6cm}|}
    \hlineB{3}
        & \bf DBpedia & \bf \begin{tabular}{@{}c@{}} {\small YahooAns.} \\ {\small (Upper) } \end{tabular} & \bf \begin{tabular}{@{}c@{}} {\small YahooAns.} \\ {\small (Lower) } \end{tabular} & \bf AGNews & \bf Yelp Reviews & \bf IMDB\\
    \hlineB{3}
    {\small\tt Base TextCNN} & 98.67 & 74.52 & 52.02 & 91.63 & 57.82 & 81.71 \\
    \hline
    {\small\tt + GROVER} & +0.05 & +1.17 & +1.80 & +0.28 & +0.16 & -0.72 \\
    \hlineB{3}
    {\small\tt + DO (p=0.1)} & 98.72 & 74.49 & 51.99 & 91.84 & 58.00 & 81.41 \\
    \hline
    {\small\tt + DO\&GROVER} & +0.01 & +1.20 & +1.95 & +0.03 & +0.07 & -0.28 \\
    \hlineB{3}
    {\small\tt + BN} & 98.50 & 74.43 & 51.36 & 91.41 & 57.58 & 81.27 \\
    \hline
    {\small\tt + BN\&GROVER} & +0.13 & +1.00 & +1.75 & +0.14 & +0.23 & -0.27 \\
    \hlineB{3}
    {\small\tt + LN} & 98.75 & 75.46 & 53.38 & 91.56 & 57.63 & 81.19 \\
    \hline
    {\small\tt + LN\&GROVER} & +0.08 & +0.90 & +1.11 & +0.21 & +0.17 & -0.05 \\
    \hlineB{3}
    {\small\tt + DO\&BN} & 98.51 & 74.90 & 51.48 & 91.18 & 57.86 & 80.81 \\
    \hline
    {\small\tt + DO\&BN\&GROVER} & +0.10 & +0.93 & +1.75 & +0.31 & +0.29 & +0.00 \\
    \hlineB{3}
    {\small\tt + DO\&LN} & 98.81 & 75.34 & 53.78 & 91.49 & 58.18 & 81.19 \\
    \hline
    {\small\tt + DO\&LN\&GROVER} & +0.07 & +1.17 & +1.09 & +0.05 & +0.12 & -0.14 \\
    \hlineB{3}
    \end{tabular}
    \end{table*}
    
    \begin{table*}[t] \centering
    \caption{List of top-20 nearest words of a cue word ({\tt love}) in initial embedding (Initial), after fine-tuned once (FineTuned), and our method GROVER in DBpedia dataset.
    The differences between initial embedding and the other methods are marked in underlined. The differences between fine-tuned once and GROVER are marked in bold. GROVER further changes the distribution of word vectors.}\label{tab:4}
    \begin{tabular}{|c||c|c|}
    \hlineB{3}
        {\small Word} & {\small Method} & Top-20 Nearest Words(Cosine Similarity)\\
    \hlineB{3}
    \multirow{7}{*}{{\small\tt love}}   & {\small Initial} &
        \begin{tabular}{@{}c@{}c@{}}
            {\small adore(.5958),hate(.5925),loved(.5786),luv(.5406),loooove(.5291),looooove(.5217),loveeee(.5177),}\\
            {\small want(.5166),loving(.5157),looove(.5071),know(.5033),loooooove(.4978),friendship(.4917),}\\
            {\small loadsss(.4895),loves(.4870),loveeeee(.4851),passion(.4797),it!i(.4727),loveee(.4692),unfeigned(.4688)}\\
        \end{tabular} \\
        \cline{2-3}
                            & {\small FineTuned} &
        \begin{tabular}{@{}c@{}c@{}}
            
            {\small adore(.5928),hate(.5858),loved(.5805),luv(.5293),loving(.5197),know(.5047),friendship(.4951),}\\
            {\small want(.4876),loves(.4839),passion(.4799),\underline{romance}(.4772),\underline{like}(.4689),\underline{affection}(.464),\underline{joy}(.4623),}\\
            {\small \underline{i}(.4572),\underline{believe}(.4526),\underline{wish}(.4522),\underline{think}(.4501),\underline{appreciate}(.4469),\underline{enjoy}(.4464)}
        \end{tabular} \\
        \cline{2-3}
                           & {\small GROVER} &
        \begin{tabular}{@{}c@{}c@{}}
            
            {\small 
            adore(.5939),loved(.5929),hate(.5646),luv(.5315),loving(.5114),passion(.4967),friendship(.4913),}\\
            {\small know(.4902),\underline{romance}(.4828),loves(.4807),want(.4714),\underline{affection}(.4684),\underline{like}(.4592),}\\
            {\small \underline{believe}(.4538),\underline{joy}(.4516),appreciate(.4495),\underline{\bf happy}(.4435),\underline{i}(.4434),\underline{\bf cherish}(.4412),\underline{\bf relationship}(.4406)}
        \end{tabular} \\
    \hlineB{3}
    \end{tabular}
    \end{table*}
    
    \begin{table*}[t] \centering
    \caption{The performance of TextCNN with GROVER according to step size, which determines how much the random maskers move to next frequently used words. Our default setting is to mask 10\% of vocabulary at once.}\label{tab:5}
    \begin{tabular}{|l||C{1.5cm}|C{1.5cm}|C{1.5cm}|C{1.5cm}|C{1.5cm}|C{1.5cm}|}
    \hlineB{3}
        & \bf DBpedia & \bf \begin{tabular}{@{}c@{}} {\small YahooAns.} \\ {\small (Upper) } \end{tabular} & \bf \begin{tabular}{@{}c@{}} {\small YahooAns.} \\ {\small (Lower) } \end{tabular} & \bf AGNews & \bf Yelp Reviews & \bf IMDB\\
    \hlineB{3}
    {\small\tt Step Size .05} & \bf 98.73 & 75.28 & 53.74 & 91.95 & 57.92 & 81.24 \\
    \hline
    {\small\tt Step Size 0.1} & 98.72 & \bf 75.69 & \bf 53.82 & 91.91 & 57.98 & 80.99 \\
    \hline
    {\small\tt Step Size 0.2} & 98.71 & 75.23 & 53.72 & \bf 92.18 & \bf 58.43 & 81.32 \\
    \hline
    {\small\tt Step Size 0.5} & \bf 98.73 & 74.57 & 52.56 & 92.01 & 58.02 & \bf 81.33 \\
    \hline 
    {\small\tt Step Size 1.0} & 98.68 & 74.31 & 51.35 & 91.53 & 58.06 & 81.43 \\
    \hlineB{3}
    \end{tabular}
    \end{table*}
    
    \begin{figure}[t] \centering
    \includegraphics[scale=0.28]{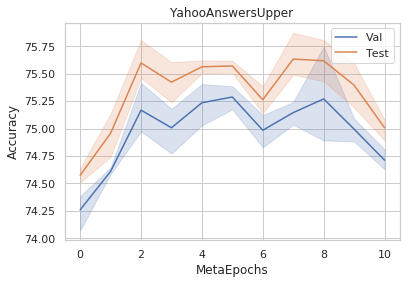}
    \includegraphics[scale=0.28]{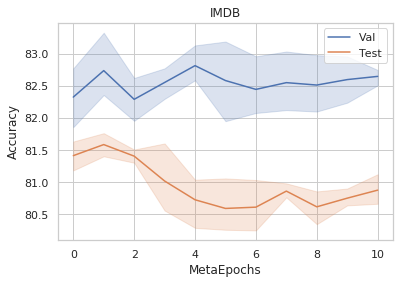}
    \caption{Training curves in YahooAnswer(Upper), and IMDB dataset, respectively. The validation performance in YahooAnswer is correlated to the test performance, whereas the validation performance in IMDB is not correlated to the test performance.}
    \label{fig:2}
    \end{figure}
    
    \begin{figure*}[t!] \centering
    \includegraphics[scale=0.32]{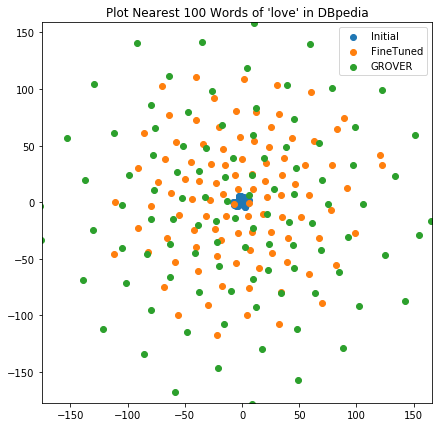} 
    \includegraphics[scale=0.32]{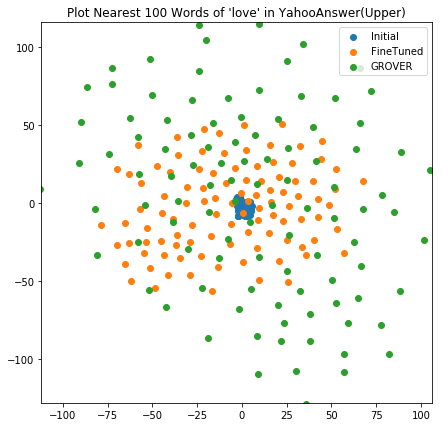} 
    \includegraphics[scale=0.32]{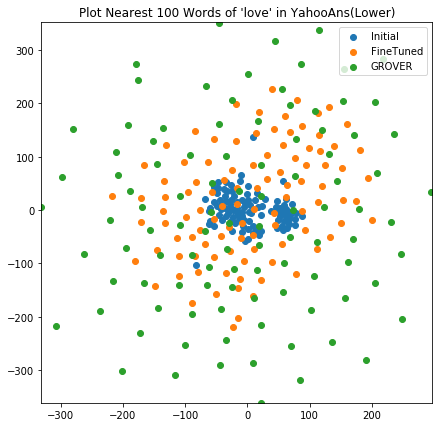} 
    \includegraphics[scale=0.32]{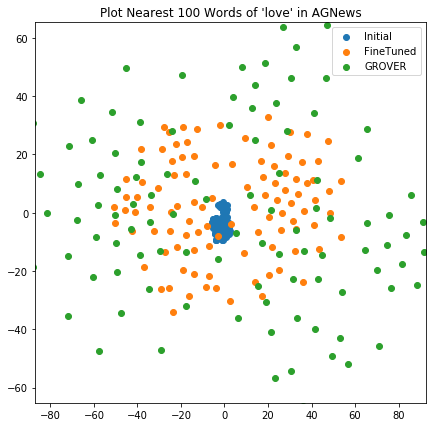} 
    \includegraphics[scale=0.32]{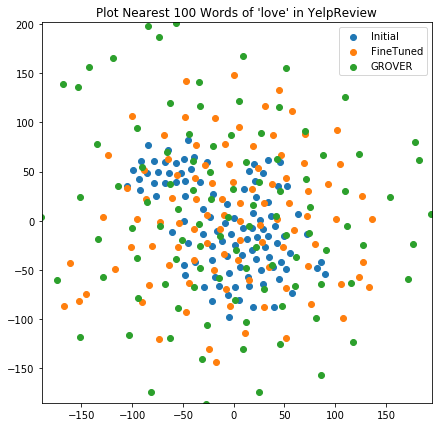} 
    \includegraphics[scale=0.32]{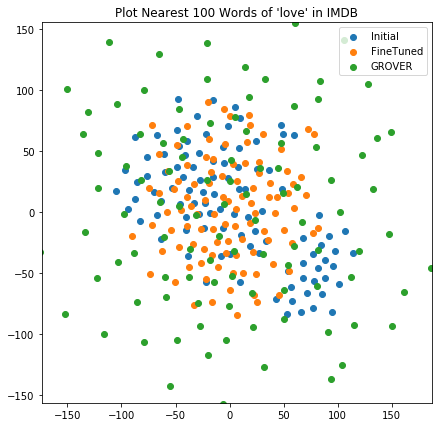} 
    \caption{Plots of nearest top-100 words of a cue word ({\tt love}) in initial embedding (Initial), after fine-tuned once (FineTuned), and our method GROVER in 5 text classification datasets (YahooAnswers dataset is used in 2 different ways).
    Note that the word vector distribution is largely changed through GROVER when compared with fine-tuned once.
    The distribution of GROVER is trained further than fine-tuned embedding.
    }
    \label{fig:3}
    \end{figure*}

\section{Results}
\subsection{Performance}\label{sec:5.1}
    We experiment with our method when initialized with 5 different pretrained word embeddings: word2vec~\cite{mikolov2013efficient} {\small GoogleNews-vectors-negative300.bin}, GloVe~\cite{pennington2014glove} {\small glove.42B.300d.txt}, fastText~\cite{bojanowski2016enriching} {\small wiki-news-300d-1M-subword.vec}, extracted token embedding from BERT~\cite{devlin2018bert}\footnote{\url{https://github.com/Kyubyong/bert-token-embeddings}} {\small bert-base-uncased.30522.768d.vec}, and extrofitted GloVe~\cite{jo2018extrofitting}. The results are presented in Table~\ref{tab:2}.\\ GROVER improves the performance on most of the datasets when even using randomly initialized word vectors. Since we train a model from scratch except for word embeddings, the result implies that we can learn better word representations through GROVER.
    However, in some case, GROVER degrades the model performance because the distribution of word vectors in pretrained embeddings is already good enough for the tasks.\\
    The model performance with regularization techniques is presented in Table~\ref{tab:3}. The result shows that the performance gain is comparable to other regularization methods except for IMDB. This might be because the number of data in IMDB is small so the validation set cannot represent the distribution of the test set. Thus, the model tries to fit the validation set through early-stopping, but the improvement in validation set does not lead to the improvement in test set. We present both the validation performance and the test performance on YahooAnswer(Upper) and IMDB, where GROVER shows good and bad performance, respectively (see Figure~\ref{fig:2}). The results show that the improvements in validation set do not warrant the improvements in test set. Another reason might be from the degree of noises added by GROVER. The noises might disturb the features in IMDB dataset too much. The further ablation study with respect to the degree of noise will be discussed.\\
    We also present the results when our method is combined with other regularization methods in Table~\ref{tab:3}. The result shows that GROVER positively matches with the other regularization techniques, further improving the model performance.
    
\subsection{Word Representations}
    Next, we analyze the embeddings further fine-tuned through GROVER.\\
    We extract the word representations updated on each datasets and plots top-100 nearest words. We visualize the distribution of word representation using t-SNE~\cite{maaten2008visualizing}, as presented in Figure~\ref{fig:3}. We can see that the distribution of frequently used words is trained further than fine-tuned embedding. These results show that our method can change the word vector distribution to be specialized further.\\
    We present the list of top-20 nearest words of a cue word in Table~\ref{tab:4}. We can also observe that the word vectors are further fine-tuned. Moreover, we can find other similar words that are not shown in fine-tuned once embedding.
    
    \begin{table*}[ht!] \centering
    \caption{The performance of TextCNN with GROVER according to the range of random noises. Our default setting is 1, which means the random values are in the range between -1 and 1.}\label{tab:6}
    \begin{tabular}{|l||C{1.5cm}|C{1.5cm}|C{1.5cm}|C{1.5cm}|C{1.5cm}|C{1.5cm}|}
    \hlineB{3}
        & \bf DBpedia & \bf \begin{tabular}{@{}c@{}} {\small YahooAns.} \\ {\small (Upper) } \end{tabular} & \bf \begin{tabular}{@{}c@{}} {\small YahooAns.} \\ {\small (Lower) } \end{tabular} & \bf AGNews & \bf Yelp Reviews & \bf IMDB\\
    \hlineB{3}
    {\small\tt Noise Range 0.1} & 98.72 & 75.31 & \bf 54.13 & 92.03 & 57.10 & 80.76 \\
    \hline
    {\small\tt Noise Range 0.5} & \bf 98.74 & 75.26 & 53.42 & \bf 92.04 & 57.59 & \bf 81.64 \\
    \hline
    {\small\tt Noise Range 1.0} & 98.72 & \bf 75.69 & 53.82 & 91.91 & \bf 57.98 & 80.99 \\
    \hline
    {\small\tt Noise Range 2.0} & 98.57 & 75.49 & 53.84 & 91.76 & 57.29 & 80.85 \\
    \hline
    {\small\tt Noise Range 10.} & 97.83 & 62.62 & 38.54 & 87.24 & 54.51 & 72.34 \\
    \hlineB{3}
    \end{tabular}
    \end{table*}
    
    \begin{table*}[ht!] \centering
    \caption{The performance of TextCNN classifiers with GROVER according to the policy of gradualness. The default method is to increase the number of maskers when the validation performance decrease.}\label{tab:7}
    \begin{tabular}{|l||C{1.5cm}|C{1.5cm}|C{1.5cm}|C{1.5cm}|C{1.5cm}|C{1.5cm}|}
    \hlineB{3}
        & \bf DBpedia & \bf \begin{tabular}{@{}c@{}} {\small YahooAns.} \\ {\small (Upper) } \end{tabular} & \bf \begin{tabular}{@{}c@{}} {\small YahooAns.} \\ {\small (Lower) } \end{tabular} & \bf AGNews & \bf Yelp Reviews & \bf IMDB\\
    \hlineB{3}
    {\small\tt Proposed GROVER} & 98.72 & 75.69 & \bf 53.82 & \bf 91.91 & \bf 57.98 & \bf 80.99 \\
    \hlineB{3}
    {\small\tt No Gradualness} & \bf 98.73 & 75.18 & 53.59 & 91.87 & 57.63 & 80.24 
    \\
    \hline
    {\small\tt Reversed Grad.} & 98.72 & \bf 75.75 & 53.48 & 91.79 & 57.64 & 80.40 \\
    \hline
    {\small\tt Both Grad.} & 98.71 & 75.66 & 53.61 & 91.87 & 56.88 & 80.98 \\
    \hlineB{3}
    \end{tabular}
    \end{table*}
    
\section{Ablation Studies}\label{sec:6}
\noindent{\bf Degree of Noises (Step Size \& Noise Range).}\label{sec:6.3}
    The degree of noises added by GROVER is an important factor in that some noises should be small enough to be corrected during the training processes, while other noises should be large enough to change the word vector distribution. We first change the step size of how much random maskers move in every training processes. The bigger step size becomes, the more words are masked in a training process, so the degree of noises increases. Likewise, the degree of noises decreases, as the step size becomes small. The effect of step size is presented in Table~\ref{tab:5}. Moreover, the range of random values that we fill in the maskers also affects the degree of noises. We present the performance according to the noise range in Table~\ref{tab:6}.\\
    We find that the degree of noises, which is controlled by the step size and the noise ranges, should be carefully chosen depending on the dataset. However, with an appropriate degree of noise, GROVER always performs well.\\
\noindent{\bf Gradualness. }
    Our proposed method is to increase the number of maskers when the validation performance decrease in order to make noises on the previous words again. Otherwise, we simply move to the next frequently used words. We change the gradualness that (1) do not have gradualness, (2) have gradualness only when the validation performance increase, which is reverse to our proposed method, and (3) have gradualness both when the validation performance increase and decrease. The result is presented in Table~\ref{tab:7}. Although our proposed approach shows the best, the policy of gradualness is a hyperparameter in that the performance gap is not much different.
    
\section{Discussion}
    {\bf As a regularization technique.} GROVER prevents the model from overfitting to input features (word vectors) by slightly modifying a portion of the word vectors with random noises. Our method shows performance gain on most of text classification datasets. Moreover, GROVER can be adapted to other regularization techniques, bringing further improvements.\\
    {\bf As a representation learning.} Recent research related to contextual representation~\cite{peters2018deep,devlin2018bert} largely improve the model performance, but the pretrained embedding models require lots of additional computational costs and data resources. With our method, we believe that such contextual information can be learned through the word vector updating in iterative training processes.\\
    Furthermore, our final representations are learned from a given training set only and do not require additional embedding model. That is, the representations are specialized into the model architecture to solve the given task. Although our method requires additional training time, the performance gain by GROVER is useful when the pretrained word vectors are not suitable for given tasks (e.g., random), or when collecting additional data is hard.
    
\section{Conclusion}
    We propose GROVER, which adds random noises to word embeddings to change its word vector distribution and regularize a model. Through the re-training process, we can mitigate some noises to be compensated and utilize other noises to learn better representations. In the experiments, GROVER regularizes the model while the model with GROVER incrementally fits to the validation set through early-stopping. We expect that our method can be utilized to improve model performances and to get better representation specialized on a given task.

\bibliographystyle{aaai}
\bibliography{aaai20}



    

\end{document}